\documentclass[10pt,twocolumn,letterpaper,hyphens]{article}
\usepackage[table]{xcolor}
\usepackage[utf8]{inputenc} % allow utf-8 input
\usepackage[T1]{fontenc}    % use 8-bit T1 fonts
\usepackage{url}            % simple URL typesetting
\usepackage{booktabs}       % professional-quality tables
\usepackage{amsfonts}       % blackboard math symbols
\usepackage{nicefrac}       % compact symbols for 1/2, etc.
\usepackage{microtype}      % microtypography
\usepackage{xcolor}         % colors
\usepackage{amsmath} 
\usepackage{graphicx}
\usepackage{wrapfig}
\usepackage{multirow} 
\usepackage[accsupp]{axessibility}
\usepackage{float} 
\usepackage{algorithm}
\usepackage{algpseudocode}  % 或者 \usepackage{algorithmic}

\usepackage{booktabs}  % 为表格添加专业线条
\usepackage{array}     % 增强表格功能
\usepackage{wrapfig} 
\usepackage{booktabs}
\usepackage{multirow}
\usepackage{makecell}
\usepackage{graphicx}
\usepackage{adjustbox}
\usepackage[most]{tcolorbox}
\newcommand{\myPara}[1]{\vspace{5pt}\noindent$\bullet$~\textbf{#1} \ }
\usepackage{newtxtext}
\usepackage{newtxmath}
\newcommand{\withdelta}[2]{%
  \pgfmathsetmacro{\diff}{#1-#2}%
  \ifdim \diff pt>0pt
    #1\ (\,+\pgfmathprintnumber[fixed,precision=2]{\diff}\,)
  \else
    #1\ (\,\pgfmathprintnumber[fixed,precision=2]{\diff}\,)
  \fi
}

\usepackage{cvpr}              % To produce the CAMERA-READY version
\definecolor{cvprblue}{rgb}{0.21,0.49,0.74}
\usepackage[pagebackref,breaklinks,colorlinks,allcolors=cvprblue]{hyperref}

\def\paperID{4970} % *** Enter the Paper ID here
\def\confName{CVPR}
\def\confYear{2026}

\title{Extending One-Step Image Generation from Class Labels to Text \\
via Discriminative Text Representation}

\makeatletter
\def\@author{
    Chenxi Zhao$^{1}$\thanks{Work done during the internship at AMAP.}, Chen Zhu$^{2}$, Xiaokun Feng$^{2}$, Aiming Hao$^{2}$, Jiashu Zhu$^{2}$, \\
    Jiachen Lei$^{2}$, Jiahong Wu$^{2}$\thanks{Project Leader.}, Xiangxiang Chu$^{2}$, Jufeng Yang$^{1}$\thanks{Corresponding author.}\\
    {$^1$ College of Computer Science, Nankai University}
    {$^2$ AMAP, Alibaba Group} 
 \\
    
    {\small \texttt{zhaochenxi@mail.nankai.edu.cn} \qquad \texttt{hongxi.wjh@alibaba-inc.com}}
}
\makeatother

\begin{document}

\twocolumn[{%
\renewcommand\twocolumn[2][]{#1}%
\maketitle
\thispagestyle{empty}
\vspace{-14mm}
\begin{center}
   \centering
   \includegraphics[width=1.0\linewidth]{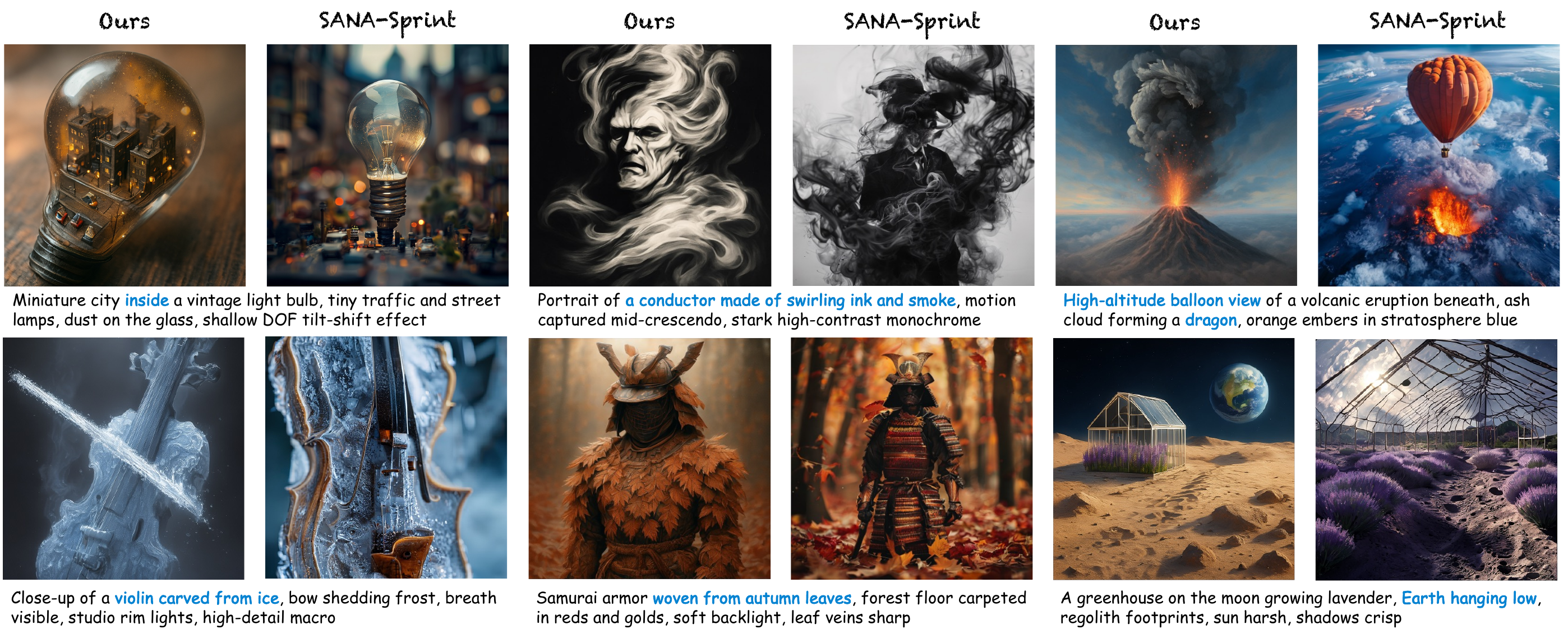}
   \vspace{-2em}
   \captionof{figure}{\textbf{Visual comparison of Our Model and SANA-Sprint in 4-step inference on challenging text.} Our model achieves superior image quality and instruction following. The blue text denotes examples where SANA-Sprint fails.}
\end{center}

}]

{
  \renewcommand{\thefootnote}{\fnsymbol{footnote}} 
  \footnotetext[1]{Work done during the internship at AMAP. \quad 
  $^\dagger$Project Leader.}
  \footnotetext[3]{Corresponding author.}
}

\begin{abstract}
Few-step generation has been a long-standing goal, with recent one-step generation methods exemplified by MeanFlow achieving remarkable results. Existing research on MeanFlow primarily focuses on class-to-image generation. However, an intuitive yet unexplored direction is to extend the condition from fixed class labels to flexible text inputs, enabling richer content creation. Compared to the limited class labels, text conditions pose greater challenges to the model’s understanding capability, necessitating the effective integration of powerful text encoders into the MeanFlow framework. Surprisingly, although incorporating text conditions appears straightforward, we find that integrating powerful LLM-based text encoders using conventional training strategies results in unsatisfactory performance. To uncover the underlying cause, we conduct detailed analyses and reveal that, due to the extremely limited number of refinement steps in the MeanFlow generation, such as only one step, the text feature representations are required to possess sufficiently high discriminability. This also explains why discrete and easily distinguishable class features perform well within the MeanFlow framework. Guided by these insights, we leverage a powerful LLM-based text encoder validated to possess the required semantic properties and adapt the MeanFlow generation process to this framework, resulting in efficient text‑conditioned synthesis for the first time. Furthermore, we validate our approach on the widely used diffusion model, demonstrating significant generation performance improvements. We hope this work provides a general and practical reference for future research on text-conditioned MeanFlow generation. The code is available at \url{https://github.com/AMAP-ML/EMF}.
\end{abstract}    
\vspace{-1em}
\section{Introduction}

Generative models, exemplified by diffusion models \cite{ho2020denoising,song2020denoising} and flow matching \cite{lipmanflow,esser2024scaling}, have achieved remarkable success in image content creation.
Since generating high-quality images typically requires many denoising iterations, few-step generation \cite{song2023consistency,heek2024multistep,sabour2025align,boffi2024flow} aims to reduce denoising steps to improve efficiency, becoming an active research direction.
As a representative and promising approach, flow map methods \cite{sabour2025align,boffi2024flow} model the average velocity between two time steps, enabling efficient one-step generation.
In particular, MeanFlow \cite{geng2025mean}, a principled extension of flow matching, shows that flow maps can achieve performance comparable to standard models.

The acceleration potential demonstrated by MeanFlow has garnered widespread interest in subsequent research \cite{zhang2025alphaflow,guo2025splitmeanflow,lee2025decoupled}. 
% For example, XXX (2 follw-up works). 
Although these studies improve MeanFlow from various perspectives, their experiments primarily focus on class label conditioned image generation in the ImageNet setting \cite{deng2009imagenet}.
To enable richer and more diverse content creation, an intuitive yet unexplored direction is to extend the conditioning from fixed class labels to flexible text inputs.
Compared to the limited class labels, text conditions impose greater challenges on the generative model’s semantic understanding capabilities. 
Adapting to text conditioning thus necessitates the effective integration of powerful text encoders into the MeanFlow framework.

To enhance text understanding and instruction-following capabilities, modern text-to-image(T2I) generation models, such as SANA-1.5 \cite{xie2024sana}, commonly replace the CLIP \cite{radford2021learning} or T5 \cite{raffel2020exploring} encoder with large language models (LLMs) \cite{team2024qwen2,team2024gemma2}.
Following their practice, we attempt to integrate LLM-based text encoders into the MeanFlow framework to achieve one-step T2I generation.
Surprisingly, we find that directly applying the widely used diffusion training loss to adapt LLM-based text encoders with diffusion models fails to yield satisfactory results.
This motivates us to conduct detailed analyses to uncover the underlying cause.

The well-known stability issue of the JVP term has been repeatedly identified as the primary bottleneck in scaling consistency-based methods to large-scale applications such as T2I~\cite{chen2025sana,lu2024simplifying,zheng2025large}, so directly applying Mean Flow to T2I tasks is not an easy task.
By contrasting our failed experiments with those in which MeanFlow succeeds, we offer two observations. 
First, fine‑tuning MeanFlow on a pretrained model is substantially easier than training from scratch on DiT: a pretrained model already encodes a velocity field, so MeanFlow mainly needs to map instantaneous to average velocity~\cite{geng2024consistency,lu2024simplifying,song2023improved}. Yet this presumed advantage is doubtful, since even large, state‑of‑the‑art text‑to‑image models—which are more expressive than DiT—also struggle to learn the average velocity, calling into question the practical benefit of the pretrained velocity field for MeanFlow.
Second, a recent line of work \cite{lei2025advancing,zheng2025diffusion,yu2024representation,shi2025latent,wu2025representation,chu2025usp,xie2025reconstruction} investigates representation learning for image generation, aiming to enhance class separability and improve generation quality through stronger visual representations. Unlike class-conditional settings—where supervision is relatively clean and unambiguous and most studies center on diffusion models—T2I relies on complex textual conditions whose semantics must be carefully parsed and precisely grounded in the visual space. As a result, making MeanFlow effective hinges on prioritizing the quality of the text encoder. Yet the true role of text encoders in visual generation remains insufficiently understood.

To further validate our hypothesis, we conducted the following analyses. To probe the instantaneous velocity in the generative velocity field, we examined how reducing the number of denoising iterations affects different text representations. Specifically, we evaluated standard generative models equipped with various text encoders under limited-iteration settings. The results reveal substantial differences across text representations in their ability to preserve semantic fidelity when the number of steps is constrained. This indicates that, although some models achieve strong final performance, their underlying velocity fields may be of low quality and are only corrected through multiple denoising steps~\cite{xie2025sana}.
Through targeted analyses of the text encoders, we distilled two core insights:1) High-quality text representations are required to exhibit strong semantic discriminability, effectively capturing subtle differences among semantically similar texts; 2) They  also need to possess good semantic disentanglement, clearly reflecting the distinct semantic components within the text. These two properties help reduce the difficulty of semantic discrimination under limited denoising steps, thereby improving the semantic fidelity of the generative model.
We hypothesize that these characteristics alleviate the semantic discrimination burden faced by generative models under a limited number of denoising steps, thereby making them better suited to the MeanFlow framework.

Building on these insights, we, for the first time, enable MeanFlow to be effectively applied to T2I generation. Specifically, we validate the proposed method on the recently popular diffusion model BLIP3o-NEXT, achieving significant improvements across multiple evaluation benchmarks while demonstrating the scalability of our approach. We hope this work provides a general and practical reference for future research on text-conditioned MeanFlow generation.

In summary, our contributions are as follows:
\begin{itemize}
    \item To the best of our knowledge, we are the first to explore and realize the extension of conditioning in MeanFlow-based one-step generation from fixed class labels to flexible text inputs, enabling rich and efficient image generation.
    \item Integrating powerful LLM-based text encoders into the Mean Flow framework, we find that under a limited number of iterations, different textual representations yield velocity fields of varying quality, which in turn induce pronounced differences in semantic fidelity. Furthermore, we systematically analyze the key properties of high-quality textual representations—\textbf{discriminability} and \textbf{disentanglement}.
    \item Experiments on BLIP3o-NEXT validate our design, and our one-step T2I model, \textbf{EMF} (Extending MeanFlow to T2I), achieves competitive results on standard benchmarks.

\end{itemize}

\section{Related Work}

\subsection{Text to Image Generation}
The field of video generation has witnessed significant advancements over the past year. These improvements stem from multi-faceted innovations: architectural transitions from U-Net \cite{ronneberger2015u} to DiT \cite{peebles2023scalable,esser2024scaling}, denoising paradigm shifts from diffusion \cite{ho2020denoising,song2020denoising} to flow matching  \cite{lipmanflow,esser2024scaling,chen2025s2guidancestochasticselfguidance,chen2025taming} optimization, and the evolution of text encoders from early text foundation models \cite{radford2021learning,raffel2020exploring} to LLMs \cite{bai2023qwen,team2024qwen2,yang2025qwen3,team2024gemma,team2024gemma2,team2025gemma3}.
Representative works such as the Stable Diffusion \cite{rombach2022high,podell2023sdxl,esser2024scaling} and PixArt \cite{chen2024pixartalpha,chen2024pixart,chen2024pixartsigma} series have continuously improved image generation capabilities.
Recent large-scale models like FLUX \cite{flux2024,lan2025flux}, Nano Banana \cite{google_nanobanana_2025}, Qwen-Image \cite{wu2025qwen}, and HunyuanImage 3.0 \cite{cao2025hunyuanimage} have demonstrated the ability to synthesize complex content and accurately edit images.
To enhance semantic understanding and instruction following abilities of generative models, models such as Playground v3 \cite{liu2024playground}, SANA-1.5 \cite{xie2024sana}, and BLIP3o-NEXT \cite{chen2025blip3o} focus on integrating LLMs \cite{team2024qwen2,team2024gemma2} effectively into the generation framework.
Meanwhile, given that high-quality image synthesis typically requires multiple denoising iterations, reducing the number of denoising steps to improve generation efficiency has become another important research direction.

\subsection{Few-step Generation}
Although diffusion models achieve excellent generation quality, their iterative sampling process incurs high computational cost. Considerable efforts have been devoted to accelerating sampling to fewer or even one step.
A representative approach is Consistency Model \cite{song2023consistency,heek2024multistep,lu2024simplifying}, which enforces self-consistency by requiring predictions remain invariant under repeated model application or temporal interpolation across varying noise levels.
Such constraints promote coherent and predictable generation trajectories, enabling accurate approximation with substantially fewer steps. Despite extensive studies on consistency models \cite{luo2023latent,geng2024consistency,song2023improved,yang2024consistency}, these methods are generally heuristic and introduced as external regularizers without rigorous theoretical foundations \cite{guo2025splitmeanflow}.
Recent works propose learning a flow map \cite{sabour2025align,boffi2024flow} between two time steps to accelerate inference by reducing discretization error.
In particular, MeanFlow \cite{geng2025mean} presents a principled framework for one-step generation, introducing average velocity as the ratio of displacement over a time interval. Unlike Flow Matching \cite{lipmanflow,esser2024scaling} that models instantaneous velocity per time step, MeanFlow rigorously derives the relation between average and instantaneous velocities and designs a theoretically grounded training objective accordingly.
Furthermore, MeanFlow achieves one-step generation performance comparable to standard multi-step models, attracting numerous follow-up improvements \cite{zhang2025alphaflow,guo2025splitmeanflow,lee2025decoupled}.
However, existing MeanFlow based studies primarily focus on class label-conditioned image generation. This work systematically explores and implements extending conditioning from fixed class labels to flexible text inputs.

\section{Method}
\subsection{Preliminary}

\paragraph{MeanFlow.}
To avoid the costly ODE integration in standard flow matching inference, MeanFlow learns a flow map $u_{\theta}(z_t,t,r)$ that directly predicts the transition from $z_t$ at time $t$ to $z_r$ at time $r$. The transition is defined as
\begin{equation}
    z_r = z_t + (r-t)u_{\theta}(z_t,t,r), \quad r>t.
\end{equation}
For the true ODE trajectory, the ideal flow map corresponds to the average velocity over $[t,r]$. However, computing this quantity requires access to the full trajectory and is therefore expensive in practice. MeanFlow instead derives a self-consistent target by differentiating the transition equation along the trajectory, which gives
\begin{equation}
    u_{\theta}(z_t,t,r)=v(z_t,t)+(r-t)\frac{d}{dt}u_{\theta}(z_t,t,r).
\end{equation}
Here, the total derivative is computed as $\frac{d}{dt}u_{\theta}=\partial_t u_{\theta}+(\nabla_{z_t}u_{\theta})v(z_t,t)$, which can be efficiently implemented via JVP. Based on this relation, the target is defined as $\tilde{u}(z_t,t,r)=v(z_t,t)+(r-t)\frac{d}{dt}u_{\theta}(z_t,t,r)$, and the model is trained with
\begin{equation}
    \mathcal{L}_{\mathrm{MF}}(\theta)
    =
    \mathbb{E}_{t,z_t,r}
    \left[
    \left\|
    u_{\theta}(z_t,t,r)-\mathrm{sg}(\tilde{u}(z_t,t,r))
    \right\|^2
    \right],
\end{equation}
where $\mathrm{sg}(\cdot)$ denotes the stop-gradient operator for stable optimization.

\subsection{Different Text Representations Show Distinct Few-Step Generation Potentials}
\begin{figure*}
    \centering
    \includegraphics[width=1.0\linewidth]{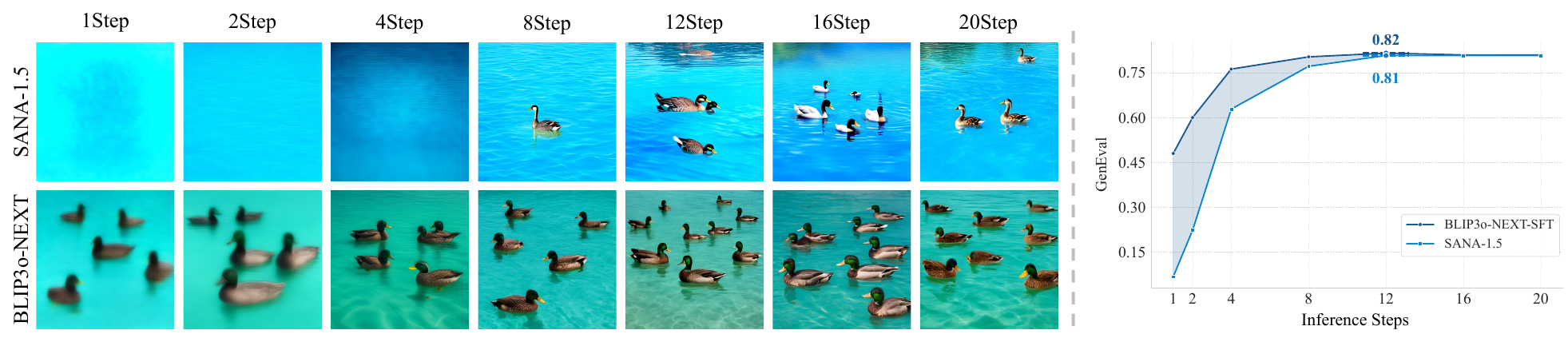}
    \vspace{-2em}
    % \caption{\textbf{Performance gap in few-step generation}: For “Ducks float leisurely on vibrant, clear blue water,” SANA-1.5 misses the subject “ducks” in early steps, while BLIP3o-NEXT preserves it, showing greater robustness and a more accurate velocity-field direction.}
    \caption{Left: \textbf{Performance gap in few-step generation} — For "Ducks float leisurely on vibrant, clear blue water.", SANA-1.5 misses the subject "ducks" in early steps, while BLIP3o-NEXT preserves it, yielding greater robustness and a more accurate velocity-field direction. Right: Under few-step sampling, \textbf{BLIP3o-NEXT consistently outperforms SANA-1.5 on GenEval}. BLIP3o-NEXT shows stronger subject preservation and downstream metric gains in few-step regimes.}
    \vspace{-1em}
    
    \label{fig:duck}
\end{figure*}
Existing studies on MeanFlow have achieved significant progress in class label-conditioned image generation tasks. This work attempts to extend MeanFlow to T2I generation, aiming to support richer and more diverse content creation.
To enable semantic understanding and instruction following for flexible text inputs, recent mainstream T2I models have gradually replaced earlier foundational text encoders (such as CLIP \cite{radford2021learning} and T5 \cite{raffel2020exploring}) with powerful LLM-based text encoders.
Following this trend, we attempt to effectively adapt LLM-based text encoders to the MeanFlow framework.

Reducing the number of generation steps limits a model’s refinement capacity \cite{wang2025transition,xie2025sana}. From the perspective of the velocity field, fewer steps are equivalent to taking a larger step along the instantaneous velocity at each time step, thereby reducing opportunities for gradual corrections to trajectory details and semantic boundaries. To examine how fewer steps affect different text representations, we evaluate two standard generative models (SANA-1.5 \cite{xie2024sana} and BLIP3o-NEXT \cite{chen2025blip3o}) under constrained-iteration settings. These models share the same diffusion backbone but employ distinct text encoders. As shown in Fig.~\ref{fig:duck}, even when the total number of steps is drastically reduced to 1, BLIP3o-NEXT maintains basic semantic integrity, whereas SANA-1.5 exhibits a substantial loss of semantic fidelity under few-step settings. The results in Fig.~\ref{fig:duck} further indicate that different text representations display varying robustness to velocity-field integration errors induced by step reduction. Evidently, the text representation associated with BLIP3o-NEXT demonstrates higher potential and quality for few-step generation; its ability to preserve basic semantic integrity even in the one-step regime suggests that the direction of BLIP3o-NEXT’s velocity field is more correct and better aligned with the target semantics. Subsequent experiments also confirm that this representation is better suited for MeanFlow-based one-step generation.

\subsection{High-Quality Text Representations Exhibit Discriminability and Disentanglement}
\label{Sec: High-Quality Text Representations Exhibit Discriminability and Disentanglement}
\begin{figure*}
    \centering
    \includegraphics[width=1.0\linewidth]{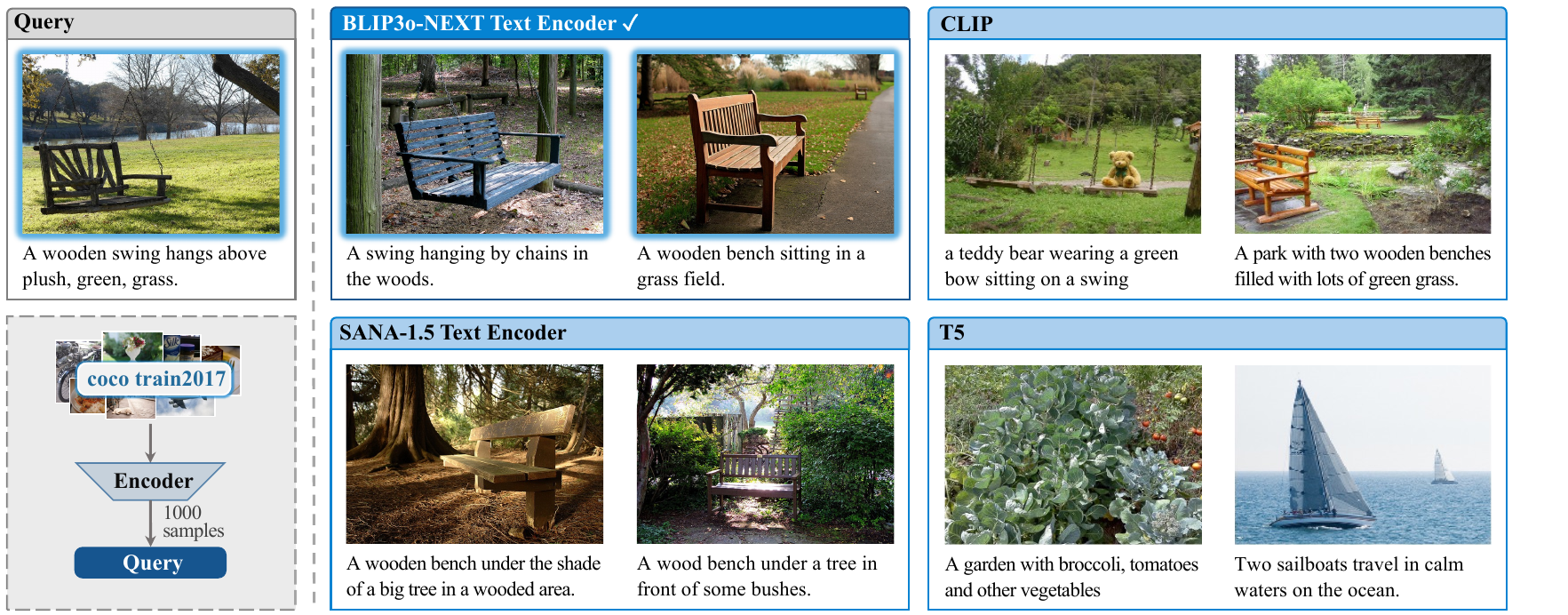}
    \caption{On the COCO2017 train set~\cite{lin2014microsoft}, we encode query prompts using different text encoders, retrieve the Top-2 similar texts, and visualize their corresponding images.  \textbf{Among them, the images retrieved using the BLIP3o-NEXT text encoder are the most similar}. This indicates that the distributions of its text and image representations are closely aligned, exhibiting strong discriminability.}
    \label{fig:query}
    \vspace{-1em}
\end{figure*}
During image synthesis, the textual condition directly governs the quality of the generated output. When the text encoder is inadequate, it struggles to build a proper velocity field, causing slow model convergence and often requiring several corrective steps before an image aligns with its textual description. As stated in the previous section, under a multi-step sampling setting, the model can repair the sampling denoising trajectory; thus the final outcome (quantified by GenEval) may remain unchanged, though the computational effort varies. To more precisely evaluate distinct text encoders, we examine two key properties—discriminability and disentanglement—across four encoders: the Blip3o-NEXT text encoder~\cite{chen2025blip3o}, the SANA-1.5 text encoder~\cite{xie2025sana}, Clip-vit-large-patch14~\cite{radford2021learning}, and T5-v1\_1-xxl~\cite{raffel2020exploring}.

\paragraph{Discriminability.} 
For a vision–language dataset composed of paired images and captions, an effective text encoder should generate representations that are well aligned with their corresponding image embeddings. Inspired by Wu et al.~\cite{wu2024saco}, we assess the cross-modal encoding quality through an image–text retrieval experiment. 
Specifically, on the 118k training split of COCO 2017~\cite{lin2014microsoft}, we first encode each textual prompt with the text encoder under evaluation. We then compute cosine similarities between this text embedding and the image embeddings of all 118k image–caption pairs, ranking the pairs by similarity to obtain the top-k matches. 
First, we perform mean pooling along the sequence dimension of the embeddings.
\begin{equation}
    \mathbf{h}(x)=\frac{1}{L_{seq}}\sum_{t=1}^{L_{seq}}\mathbf{e}_t^{(x)},
\end{equation}
Then we compute the cosine similarity.
\begin{equation}
    \operatorname{cos}(x,y)
=1-\frac{\mathbf{h}(x)^{\top}\mathbf{h}(y)}
        {\lVert\mathbf{h}(x)\rVert_2\,\lVert\mathbf{h}(y)\rVert_2}
\end{equation}
Fig.~\ref{fig:query} visualizes the two most similar pairs retrieved for a representative query, where the queries were drawn from 1,000 samples randomly selected from the 118k dataset.
The qualitative results reveal a clear pattern: because both the SANA-1.5 text encoder and T5 are trained exclusively on linguistic corpora and lack explicit vision–language alignment, the retrieved images exhibit low semantic relevance. In contrast, encoders such as BLIP3o-NEXT and CLIP, which are explicitly aligned on image–text pairs during pre-training, return qualitatively superior matches.
To further quantify retrieval performance, we re-encode the retrieved images with a strong vision backbone (DINOv3~\cite{simeoni2025dinov3}) and calculate the cosine similarity between these image embeddings and the embedding of the query image. The aggregated scores, reported in Tab.\ref{tab:retrieval}, provide a rigorous metric for comparing the alignment capabilities of different text encoders.
\begin{table}[h!] 
    \centering 
    \caption{DINO evaluation of image-feature similarity for text-retrieved images.} 
    \label{tab:retrieval} 
    \begin{tabular}{l|cccc} 
        \toprule 
        \textbf{Model} & \textbf{BLIP3o-NEXT} & \textbf{CLIP} & \textbf{Gemma} & \textbf{T5} \\
        \midrule 
        \textbf{Score} & 0.734 & 0.730 & 0.713 & 0.634 \\
        \bottomrule 
    \end{tabular}
\end{table}

\vspace{-1em}
\paragraph{Disentanglement.} 
Another crucial property is that the text encoder’s output should be highly disentangled. Intuitively, after encoding a complete prompt, the resulting text embedding should retain the linguistic structure of the original text—i.e., exhibit semantic disentanglement. Moreover, when we shorten the prompt via sentence reduction to form subsequences, the distances between their embeddings and that of the full prompt should remain as small as possible.

Motivated by this idea, we conducted experiments on the entire set of prompts in DPG-Bench~\cite{hu2024ella}. For each original prompt, we randomly removed portions of the text to create an ablated version. We then encoded both the original and ablated prompts with several different text encoders and recomputed the cosine distance between their embeddings. The experimental results are summarized in Tab.~\ref{tab:Disentanglement}.

The experimental results show that, compared with CLIP and T5, autoregressive text encoders trained with the next-token prediction paradigm perform better. In particular, the BLIP3o-NEXT text encoder and Gemma~\cite{team2024gemma} achieve strong results and exhibit good disentanglement.

\begin{table}[h!] 
    \centering 
    \caption{Evaluation of Text Encoder Disentanglement via Sub-sequence Similarity.} 
    \label{tab:Disentanglement} 
    \begin{tabular}{l|cccc} 
        \toprule 
        \textbf{Model} & \textbf{BLIP3o-NEXT} & \textbf{CLIP} & \textbf{Gemma} & \textbf{T5} \\
        \midrule 
        \textbf{Score} & 0.999 & 0.967 & 0.987 & 0.893 \\
        \bottomrule 
    \end{tabular}
\end{table}

\subsection{Extending MeanFlow to T2I Generation}

Building upon our evaluation in Sec.~\ref{Sec: High-Quality Text Representations Exhibit Discriminability and Disentanglement}, the BLIP3o-NEXT text encoder consistently outperforms other LLM-based text encoders in terms of both \emph{discriminability} and \emph{disentanglement}. Leveraging the strong capabilities of the BLIP3o-NEXT representation space, we propose an adaptation of the MeanFlow framework for T2I generation.

Specifically, given a pre-trained flow matching backbone conditioned on textual embeddings, we modify its architecture to explicitly support MeanFlow’s bidirectional time conditioning. In standard flow matching, a single temporal embedding layer $\phi_{\text{time}}(t)$ is used to represent the current generation time $t$. In our adaptation, we duplicate the temporal embedding parameters to yield two separate embedding layers: $\phi_{\text{interval}}(\cdot)$, $\phi_{\text{end}}(\cdot)$, encodes the interval length \(t - r\), and the segment end time \(t\), respectively.
Given a start time \(r\) and an end time \(t\), we construct the conditional temporal embedding as: $\phi_{\text{cond}}(t,r) = \phi_{\text{interval}}(t-r) + \phi_{\text{end}}(t)$.

The conditioning embedding $\phi_{\text{cond}}$ and text features $\psi_{\text{text}}(x_{\text{text}})$ jointly condition the velocity network:
\begin{equation}
    u_{\theta}\big(z_t, t, r, \psi_{\text{text}}\big) 
    = f_{\theta}\big(z_t, \phi_{\text{cond}}(t,r), \psi_{\text{text}}\big).
\end{equation}

During training, we adaptively sample timesteps $(t, r)$ from either a uniform or logit-normal distribution as follows:
\begin{equation}
    t, r \sim p(\cdot ; \mu(p), \sigma(p)), \qquad t \neq r,
\end{equation}
where $p$ is either uniform $\mathcal{U}(0, 1)$ or logit-normal, and the parameters $\mu(p)$, $\sigma(p)$ are interpolated between initial and final values according to training progress $p \in [0, 1]$. The ratio of non-equal timesteps $(t \neq r)$ is also increased adaptively throughout training. This strategy ensures balanced exposure to both short- and long-range segments, promoting stable learning of the mean velocity field.

The full training procedure for our T2I MeanFlow adaptation minimizes the standard MeanFlow objective:
\begin{equation}
    \mathcal{L}_{MF}(\theta) 
    = \mathbb{E}_{z_t,t,r} \left[ \| u_{\theta}(z_t, t, r, \psi_{\text{text}}) - \mathrm{sg}(u_{\text{tgt}}) \|^2 \right],
\end{equation}
with $u_{\text{tgt}}$ defined as:
\begin{equation}
    u_{\text{tgt}} 
    = v_{\theta}(z_t, t, \psi_{\text{text}}) 
    + (r - t) \frac{d}{dt} u_{\theta}(z_t, t, r, \psi_{\text{text}}),
\end{equation}
where $\psi_{\text{text}}$ is produced by the BLIP3o-NEXT encoder and injected as the text condition. The derivative term is computed via Jacobian–vector products as in Eq.~(14), and stop-gradient is applied to $u_{\text{tgt}}$ to stabilize training.

This adaptation extends MeanFlow to handle complex text-based conditioning in modern T2I models, enabling accurate and semantically faithful generation even in the one-step regime.

%\vspace{-0.5em}
\section{Experiment}
In this section, we provide a detailed description of our experimental setup, present the results of our method on mainstream image generation benchmarks and state-of-the-art models, and offer deeper analyses and insights.
\subsection{Implementation Details}

\paragraph{Training Recipe.} We use approximately 170,000 samples (BLIP3o-60k~\cite{chen2025blip3}, shareGPT-4o~\cite{chen2025sharegpt}, and Echo-4o~\cite{ye2025echo}) for our training. The learning rate is set to 1e-5, the batch size is 128, and the experiment runs for 150 epochs.
We conduct experiments based on the BLIP3o-NEXT model, while keeping all other experimental settings consistent with BLIP3o-NEXT.

\vspace{-1em}
\paragraph{Evaluation details.} We evaluate T2I generation on GenEval~\cite{ghosh2023geneval} and DPG-Bench~\cite{hu2024ella}.
GenEval provides a precise, attribute-focused evaluation of text–image faithfulness, while DPG-Bench emphasizes challenging long-form prompts that test instruction following and compositional robustness.
In addition, we evaluated human perceptual preferences on the HPS-v2 dataset~\cite{wu2023human}.

\subsection{Comparison with State-of-the-arts}
\begin{table*}[ht]
    \centering
    \label{tab:geneval}
    \caption{
    GenEval results for pretrained, unified, and distilled models, plus few-step comparisons of BLIP3o-NEXT vs our MeanFlow adaptation. 
    Our method attains the best distilled-model performance and rivals larger models even at 4-step sampling.
    }
    \renewcommand{\arraystretch}{0.8} 
    \footnotesize % scriptsize, 

    \begin{tabular}{lcc|cccccc|c}
    \toprule
    \textbf{Model} & \textbf{\#Params} & \textbf{Steps} & \textbf{\makecell[c]{Single\\Object}} & \textbf{\makecell[c]{Two\\Objects}} & \textbf{Counting} & \textbf{Colors} & \textbf{Position} & \textbf{\makecell[c]{Color\\Attribution}} & \textbf{Overall} \\
    \midrule
    \multicolumn{10}{c}{\textit{Pretrained Models}} \\
    \midrule
    SD3.5-L~\citep{esser2024scaling}  & 8B & 28 & 0.98 & 0.89 & 0.73 & 0.83 & 0.34 & 0.47 & 0.71 \\
    FLUX.1-dev~\citep{flux2024}  & 12B & 50 & 0.98 & 0.81 & 0.74 & 0.79 & 0.22 & 0.45 & 0.66 \\
    SANA-1.5~\citep{xie2025sana} & 4.8B & 20 & 0.99 & 0.93 & 0.86 & 0.84 & 0.59 & 0.65 & 0.81 \\
    \multirow{1}{*}{Cosmos-Predict2~\citep{nvidia2025cosmos}}&0.6B&35&1.00&0.97&0.74&0.86&0.59&0.70&0.81\\
    PixArt-$\alpha$~\citep{chen2024pixartalpha}    & 0.6B & 20  & 0.98          & 0.50        & 0.44     & 0.80    & 0.08     & 0.07              & 0.48     \\
    Lumina-Image 2.0~\citep{qin2025lumina}  & 2.6B & 50 & - & 0.87 & 0.67 & -      & - & 0.62 & 0.73     \\
    HiDream-I1-Full~\citep{cai2025hidream}     &  3B  & 50  & 1.00             & 0.98       & 0.79     & 0.91   & 0.60      & 0.72              & 0.83     \\
    Seedream 3.0~\citep{gao2025seedream}  &  /  & / & 0.99 & 0.96 & 0.91 & 0.93 & 0.47 & 0.80 &0.84 \\
    GPT Image 1 [High]~\citep{gptimage}      &  /  &   /    & 0.99          & 0.92       & 0.85     & 0.92   & 0.75     & 0.61              & 0.84     \\
    BLIP3o-NEXT~\cite{chen2025blip3o}    & 3B       &30  &0.99  &0.95&   0.88 &  0.90  & 0.92  &  0.79  & 0.91\\
    % \midrule
    \midrule
    \multicolumn{10}{c}{\textit{Unified Models}} \\
    \midrule
    MetaQuery-L~\citep{pan2025transfer}     & 3B  & 30  & -         & -       & -    & -    & -     & -              & 0.78 \\
    BLIP3-o-8B~\citep{chen2025blip3}      & 8B  & 30 & -         & -       & -    & -    & -     & -              & 0.83 \\
    OpenUni-B-512~\citep{wu2025openuni}  &  1.6B  & 30 & 0.99 & 0.91 & 0.74 & 0.90 & 0.77 & 0.73 & 0.84 \\
    Tar-7B~\citep{han2025vision}  &  9.6B  & 50 & - & 0.92 & 0.83 & 0.65 & - & - & 0.83 \\
    TBAC-UniImage-3B~\cite{xu2025tbac}  &  4.6B  & 30 & 0.99 & 0.94 & 0.77 & 0.92 & 0.83 & 0.75 & 0.87 \\
    Qwen-Image~\citep{wu2025qwen}   &  20B  & 50 & 0.99 & 0.92 & 0.89 & 0.88 &  0.76 & 0.77 & 0.87 \\
    \midrule
    \multicolumn{10}{c}{\textit{Distilled Models}} \\
    \midrule
    SDXL-LCM~\citep{luo2023latent} & 2.6B & 4 & 0.99 & 0.55 & 0.38 & 0.85 & 0.07 & 0.14 & 0.50 \\
    SDXL-Turbo~\citep{podell2023sdxl} & 2.6B & 4 & 1.00 & 0.72 & 0.49 & 0.82 & 0.11 & 0.21 & 0.56 \\
    SDXL-Lightning~\citep{lin2024sdxl} & 2.6B & 4 & 0.98 & 0.61 & 0.44 & 0.84 & 0.11 & 0.21 & 0.53 \\
    Hyper-SDXL~\citep{ren2024hyper} & 2.6B & 4 & 1.00 & 0.77 & 0.48 & 0.89 & 0.11 & 0.23 & 0.58 \\
    SDXL-DMD2~\citep{yin2024improved} & 2.6B & 4 & 1.00 & 0.76 & 0.52 & 0.88 & 0.11 & 0.24 & 0.58 \\
    SD3.5-L-Turbo~\citep{esser2024scaling} & 8B & 4 & 0.99 & 0.89 & 0.68 & 0.78 & 0.23 & 0.54 & 0.68 \\
    FLUX.1-schnell~\citep{flux2024} & 12B & 4 & 0.99 & 0.88 & 0.64 & 0.78 & 0.30 & 0.52 & 0.69 \\
    \multirow{2}{*}{SANA-Sprint~\citep{chen2025sana}} & 0.6B & 4 & 1.00 & 0.90 & 0.71 & 0.89 & 0.61 & 0.54 & 0.77 \\
     & 1.6B & 4 & 1.00 & 0.92 & 0.59 & 0.91 & 0.54 & 0.55 & 0.75 \\
    \multirow{1}{*}{rCM~\cite{zheng2025large}} & 14B & 4 & 1.00 & \bf0.98 & 0.80 & 0.86 & 0.59 & 0.73 & 0.83 \\
    % \cmidrule(lr){1-10}
    \midrule
    \multicolumn{10}{c}{\textit{BLIP3o-NEXT and Ours under Few-Step Generation}} \\
    \midrule
    
    \multirow{3}{*}{BLIP3o-NEXT~\cite{chen2025blip3o}} &3B    &1    &0.81  &0.40  &0.40  &0.56  &0.38  &0.23  &0.46 \\
    & 3B       &2  &0.92  &0.68  &0.55  &0.66  &0.60  &0.40  &0.63 \\
    & 3B       &4  &0.99  &0.93  &0.84  &0.84  &0.86  &0.70  &0.86  \\
    \cmidrule(lr){1-10}
    \multirow{3}{*}{EMF} &3B    &1   & 0.98	&0.86&	0.66	&0.69&	0.80	&0.47  &0.74\\
    & 3B       & 2   &0.99  &0.91  &0.81  &0.86  &0.86  &0.66  &0.85\\
    & 3B       &4  &\bf1.00    &0.94   &\bf0.88   &\bf0.92   &\bf0.91  &\bf0.76  &\bf0.90 \\
    \bottomrule
    \end{tabular}
    % }
\end{table*}

%  with DPG-Bench
\begin{table*}[t]
\setlength{\tabcolsep}{1pt}
\centering
    \caption{
    % Results on DPG-Bench and HPSv2.1. DPG-Bench results are reported for six evaluation dimensions.
    DPG-Bench and HPS-v2.1 results. Our MeanFlow adaptation matches BLIP3o-NEXT’s performance using far fewer sampling steps, with 4-step generation rivaling the 30-step baseline on both benchmarks.
    }
    \vspace{-0.5em}
    \renewcommand{\arraystretch}{0.8} % 缩小行高
    \setlength{\tabcolsep}{3pt}       % 减小列间距
    \footnotesize
\begin{tabular}{l|c|cccccc|ccccc}
\midrule
\multirow{2}{*}{Model} & \multirow{2}{*}{Steps} & \multicolumn{6}{|c}{DPG-Bench} & \multicolumn{5}{|c}{HPS-v2.1} \\
\cmidrule(lr){3-8} \cmidrule(lr){9-13}
 & & Global & Entity & Attribute & Relation & Other & Overall & anime & concept-art & paintings & photo & Average \\
\midrule
\multirow{4}{*}{BLIP3o-NEXT} 
 & 1  & 69.10	&73.48	&79.92&	69.09	&73.60	&57.05 & 19.77 & 17.54 & 18.23 & 18.64 & 18.54 \\
 & 2  & 79.35&	79.16	&77.71&	79.66	&82.32	&67.38  & 23.51 & 21.70 & 22.78 & 21.82 & 22.45 \\
 & 4  & 85.99	&85.04	&87.78	&86.44	&88.53&	78.15 & 28.13 & 26.22 & 27.18 & 26.30 & 26.96 \\
 & 30 &  88.55	&86.82&	90.14	&88.01&	86.21	&82.05 & 30.27 & 29.15 & 28.99 & 29.26 & 29.42 \\
\midrule
\multirow{3}{*}{EMF} 
 & 1  & 85.24 & 85.85 & 85.19 & 82.37 & 82.65 &  \textbf{77.36 (+20.31)} & 26.64 & 25.60 & 25.53 & 25.32 & \textbf{25.77 (+7.23)} \\
 & 2  & 85.63 & 88.15 & 85.96 & 85.69 & 86.20 & \textbf{79.44 (+12.06)} & 28.02 & 26.83 & 27.03 & 26.96 & \textbf{27.21 (+4.76)} \\
 & 4  & 88.01 & 87.27 & 88.24 & 88.78 & 87.68 & \textbf{81.20 (+3.05)} & 30.03 & 29.02 & 29.09 & 28.86 & \textbf{29.25 (+2.29)} \\
\midrule
\end{tabular}
% \vspace{-1.5em}
\label{tab:dpg_hps}
\end{table*}

In our GenEval tests, the model reached a score of 0.90 with just 4 sampling steps, nearly matching BLIP3o-NEXT’s 0.91, and outperforming nearly all other pretrained models, which usually require more than 20 steps. In addition, we surpassed every distilled model, which typically needs one or more teacher models during training, whereas our approach continues training from a single set of pretrained weights and still achieves high-quality generation in very few steps. Fig.~\ref{fig:placeholder} further illustrates qualitative examples from our approach, demonstrating high semantic fidelity and visual detail under 4-step generation.

As Tab.~\ref{tab:dpg_hps} shows, on the more challenging DPG-Bench our model maintains performance close to BLIP3o-NEXT, and across HPSv2 it yields substantial gains in human-preference alignment over BLIP3o-NEXT’s few-step sampling, matching the performance attained with 30 sampling steps.More experimental comparisons on DPG-Bench are provided in the appendix.

\subsection{Ablation of Sampling Steps in T2I MeanFlow}

\begin{figure}
    \centering
    \includegraphics[width=1.0\linewidth]{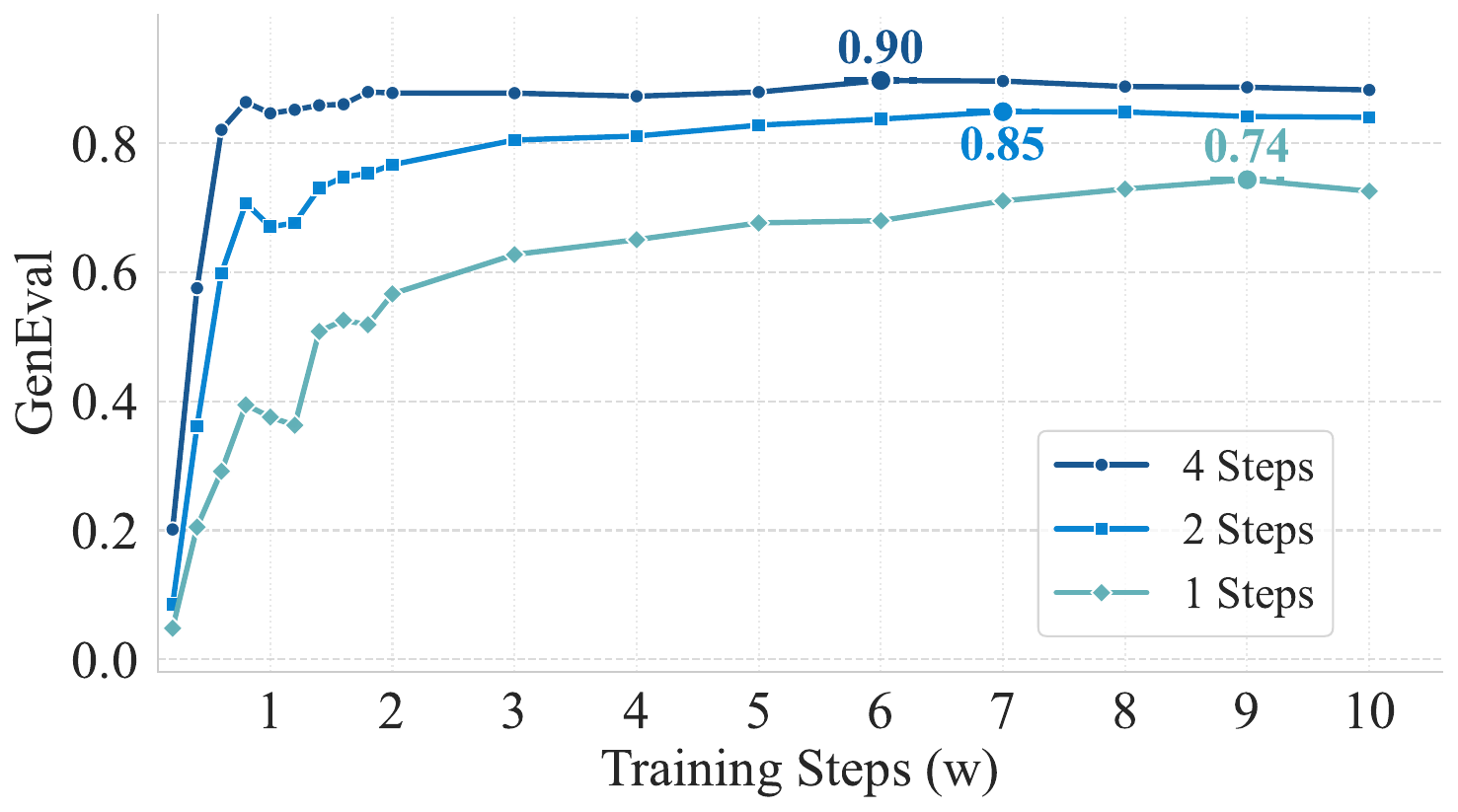}
    \caption{Ablation study of sampling steps in T2I MeanFlow. Strong 4-step performance is reached at \textasciitilde 1w training steps, while fewer steps need more training.}
    \vspace{-1em}
    \label{fig:convergence_curve}
\end{figure}

\begin{figure*}
    \centering
    \includegraphics[width=1.0\linewidth]{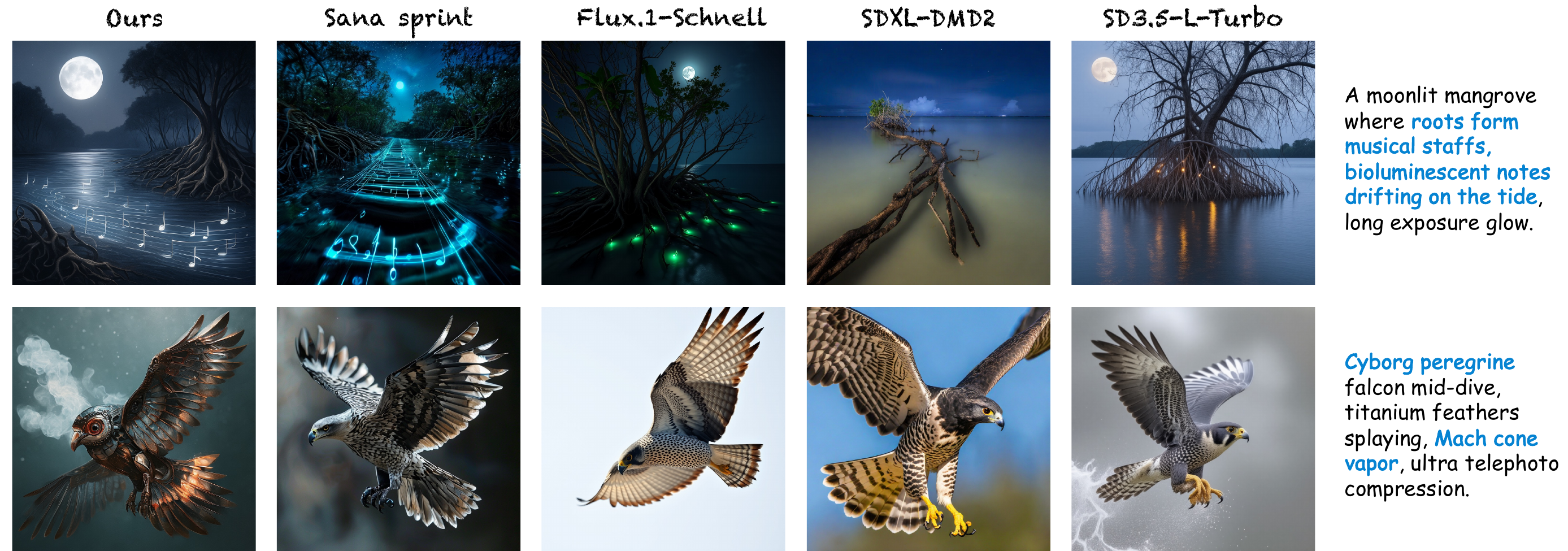}
    \caption{4-step sampling comparison of our method with existing distilled models. Our method achieves superior semantic fidelity and visual detail while closely matching complex text prompts. The blue text denotes examples where other models fail.}
    \vspace{-1em}
    \label{fig:placeholder}
\end{figure*}

To investigate the impact of sampling steps on generation quality, we monitor the performance of BLIP3o-NEXT throughout the training process under our MeanFlow framework. At each training checkpoint, the model is evaluated using the GenEval metric with three sampling configurations: 1-step, 2-step, and 4-step, as illustrated in Fig.~\ref{fig:convergence_curve}.

We observe that under our MeanFlow training framework, the model not only delivers rapid performance improvements but also converges stably across different sampling step settings. With 4-step sampling, high generation quality is achieved within roughly 10k training steps, reaching a GenEval score of 0.90 by 60k steps. Even in more challenging few-step scenarios, the framework remains robust: with 2-step sampling, the model attains a GenEval score of 0.85 at 70k steps, while 1-step sampling reaches 0.74 at 90k steps.

\section{Discussion}
Our experimental method achieves significant improvements over BLIP3o-NEXT. However, several questions remain to be discussed.

\myPara{Can our method scale beyond two steps?}
Prior work on consistency-distilled models demonstrates that they can directly produce strong image generations with very few steps. However, increasing the sampling steps in such models typically yields marginal gains in image quality, and in some cases even negative gains at larger steps~\cite{lee2025decoupled, wang2025transition, sabour2025align}, making it difficult to achieve a favorable trade-off between inference time and generation quality. 
In contrast, our model continues to benefit from additional sampling steps: performance rises from 0.74 at 1-step to 0.90 at 4-step, as shown in Fig.~\ref{fig:convergence_curve}. Notably, this 4-steps result already approaches the BLIP3o-NEXT baseline obtained with 30 sampling steps (0.91). 
Furthermore, when extending our model’s sampling steps to 8, the DPG-Bench score increases from 81.20 to 81.94 compared to the 4-step setting.
We attribute this to MeanFlow’s nature as a stable discretization of an underlying continuous generative flow—each added step more faithfully follows the average velocity field and reduces the approximation error.
As a result, our method scales gracefully beyond 2 steps, delivering sustained improvements in both quantitative metrics and perceptual fidelity without suffering from the saturation or degradation patterns observed in conventional consistency-distilled approaches.

\begin{table}[]
    \centering
    % \vspace{+2em}
    \caption{GenEval scores of SANA-1.5’s experiment. The encoder was additionally fine-tuned on SFT data to match the same domain, yet results show it still fails to achieve effective MeanFlow generation.}
    \label{tab:sana-experiment}
    \footnotesize % scriptsize, 
    \begin{tabular}{cccc|c}
    \toprule \makecell{\textbf{Sample}\\\textbf{Method}} &
\makecell{\textbf{Encoder-}\\\textbf{SFT}} &
\makecell{\textbf{MeanFlow}\\\textbf{Train}}
&\makecell{\textbf{Sampling} \\\textbf{Steps}}
 & \textbf{GenEval} \\
        % \hline
        \midrule
                Flow Matching & & & 20& 0.81 \\
                Flow Matching &  \checkmark & & 20& 0.85\\
        \midrule
                  MeanFlow & &\checkmark & 4&  0.50\\
                  MeanFlow & &\checkmark & 20& 0.83  \\
                 MeanFlow & \checkmark &\checkmark & 4& 0.47\\
                 MeanFlow & \checkmark &\checkmark & 20& 0.82 \\
    \bottomrule
    \end{tabular}
\end{table}

\myPara{Is the convergence speed of the MeanFlow dependent on the domain of the training data?}
Our initial attempt to apply MeanFlow fine-tuning on SANA-1.5 failed, likely due to a domain mismatch between SANA-1.5’s training data and MeanFlow’s data. To remove this confound, we re-trained SANA-1.5 text encoder with flow matching on the exact SFT data and hyperparameters used by BLIP3o-NEXT. As shown in Tab.~\ref{tab:sana-experiment}, encoder fine-tuning improved SANA-1.5’s GenEval score from 0.81 to 0.85, but an additional MeanFlow stage remained ineffective. Notably, while MeanFlow did not learn the average velocity field, the model reached similar performance with 20 sampling steps, indicating MeanFlow does not disrupt the original trajectories.

\begin{figure}
    \centering
    \includegraphics[width=1.0\linewidth]{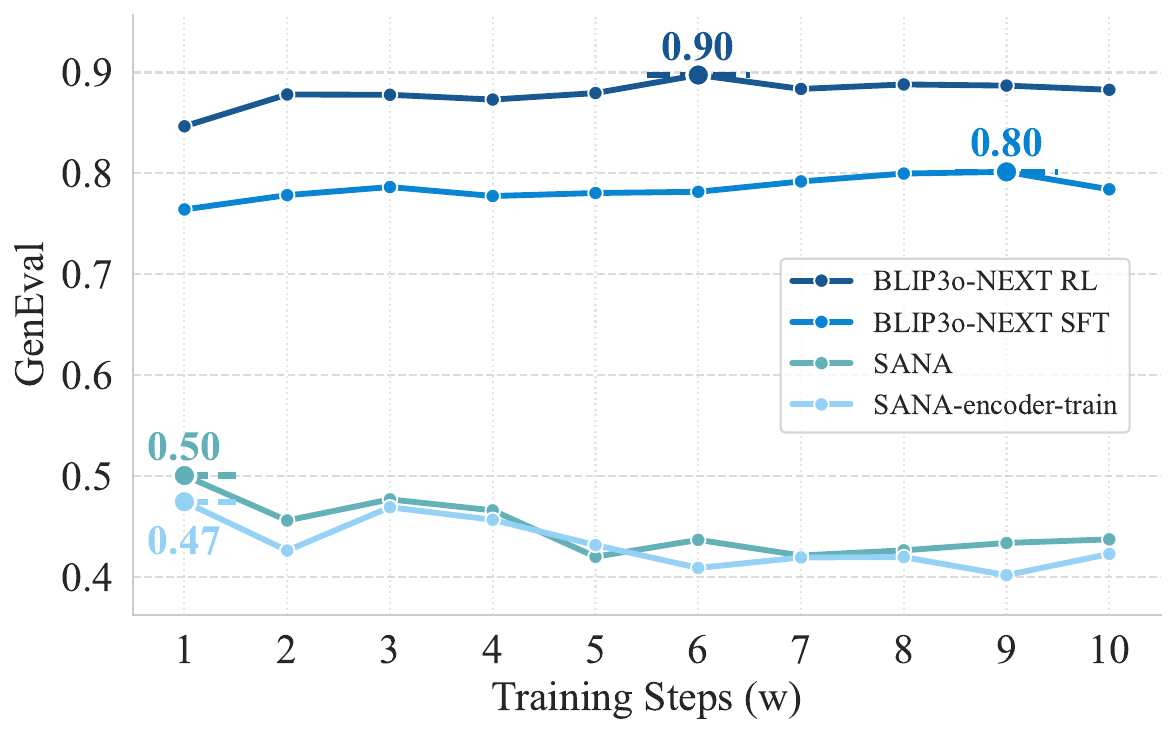}
    \caption{4-step GenEval performance of different text encoders under MeanFlow training. }
    \label{fig: variour encoders 4-step geneval}
\end{figure}

Lastly, we also trained the SFT variant of BLIP3o-NEXT with Mean Flow, and present the 4-step GenEval test results during training in Fig.~\ref{fig: variour encoders 4-step geneval}. The results again show that the SFT version of BLIP3o-NEXT converges stably, whereas SANA-1.5 exhibits training instability regardless of whether the text encoder is fine-tuned.

\section{Conclusion}
In this work, we present the first exploration and implementation of extending MeanFlow’s original class-label-conditioned one-step generation to flexible text conditioning, enabling richer and more efficient T2I synthesis. Through systematic analyses, we identify that high-quality text representations in few-step generation settings require both strong semantic discriminability and semantic disentanglement, which substantially improve semantic fidelity when only a limited number of denoising iterations are available. Guided by these insights, we adopt BLIP3o-NEXT’s powerful LLM-based text encoder—validated to possess the required semantic properties—and adapt MeanFlow on top of the BLIP3o-NEXT framework, achieving efficient text-conditioned synthesis. Empirical results validate our approach, achieving competitive one-step T2I generation with markedly improved synthesis quality. We believe this work offers valuable practical guidance and a strong reference for future research on text-conditioned MeanFlow generation.

\section{Acknowledge}
This work was supported by National Natural Science Foundation of China (No. 62532004), Shenzhen Science and Technology Program (No. JCYJ20240813114229039),  Natural Science Foundation of Tianjin (No. 24JCZXJC00040), Supercomputing Center of Nankai University.

{
    \small
    \bibliographystyle{unsrt}
    \bibliography{main}
}

\clearpage
\setcounter{page}{1}
\maketitlesupplementary

\section{Velocity Field Learning Challenges: Class-Label vs. Text Conditions}
\begin{figure}[htbp]
    \centering
    \includegraphics[width=1.0\linewidth]{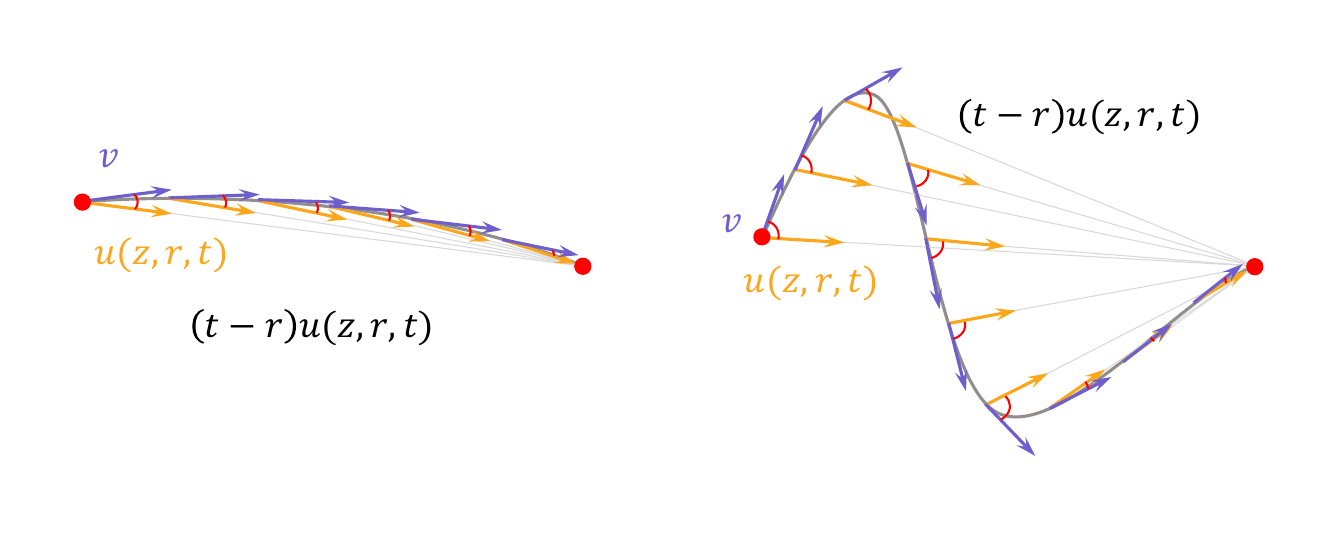}
    \caption{Denoising Trajectory Comparison. Simple class-label conditioning (left) yields a smooth path, whereas complex text conditioning (right) results in a tortuous path.}
    \label{fig:v_field}
\end{figure}
\label{sec:Velocity Field Learning Challenges: Class-Label vs. Text Conditions}

Extending MeanFlow from class-label conditioning to textual conditioning introduces fundamentally different challenges for velocity field learning.

\noindent\textbf{Representation Separability.} Class labels are discrete and well-separated in the embedding space, enabling the velocity field to maintain a stable direction. Consequently, the denoising trajectory is smooth, with the instantaneous velocity at each step closely aligning with the overall average velocity. This stability makes predicting the average velocity straightforward, ensuring high fidelity even in few-step generation. In contrast, textual embeddings form dense and continuous distributions where semantically related prompts (e.g. blue teapot vs. red teapot) occupy neighboring regions, reducing the \emph{discriminability} of the representation. This density forces the velocity field to navigate fine-grained semantic distinctions, resulting in a more tortuous trajectory. The instantaneous velocity frequently diverges from the average, leading to semantic drift and necessitating additional corrective iterations to converge on the target concept.

\begin{table*}[t!]
    \centering
    \caption{\label{tab:dpg}{Quantitative evaluation results on DPG-Bench. Our method consistently outperforms distilled few-step models of comparable scale under the same denoising step settings.}}
    \renewcommand{\arraystretch}{0.8} % 缩小行高 (默认是1.0)
    
    \setlength{\tabcolsep}{6pt}       % 减小列间距 (默认是6pt
    % \resizebox{\linewidth}{!}{
    \begin{tabular}{lcc|ccccc|c}
    \toprule
    \textbf{Model} & \textbf{\#Params} & \textbf{Steps} & \textbf{Global} & \textbf{Entity} & \textbf{Attribute} & \textbf{Relation} & \textbf{Other} & \textbf{Overall} \\
    \midrule
    \multicolumn{9}{c}{\textit{Pretrained Models}} \\
    \midrule
    PixArt-$\alpha$~\citep{chen2024pixartalpha} & 0.6B & 20 & 74.97 & 79.32 & 78.60 & 82.57 & 76.96 & 71.11 \\
    Lumina-Next~\citep{zhuo2024luminanext} & 4B & 20 & 82.82 & 88.65 & 86.44 & 80.53 & 81.82 & 74.63 \\
    Playground v2.5~\citep{li2024playground} & / & / & 83.06 & 82.59 & 81.20 & 84.08 & 83.50 & 75.47 \\
    Hunyuan-DiT~\citep{li2024hunyuandit} & 1.5B & 50 & 84.59 & 80.59 & 88.01 & 74.36 & 86.41 & 78.87 \\
    PixArt-$\Sigma$~\citep{chen2024pixartsigma} & 0.6B & 20 & 86.89 & 82.89 & 88.94 & 86.59 & 87.68 & 80.54 \\
    DALL-E 3~\citep{openai2023dalle3} & / & / & 90.97 & 89.61 & 88.39 & 90.58 & 89.83 & 83.50 \\
    FLUX.1 [Dev]~\citep{flux2024} & 12.7B & 50 & 74.35 & 90.00 & 88.96 & 90.87 & 88.33 & 83.84 \\
    SD3 Medium~\citep{esser2024scaling} & 2B & 50 & 87.90 & 91.01 & 88.83 & 80.70 & 88.68 & 84.08 \\
    HiDream-I1-Full~\citep{cai2025hidream} & 3B & 50 & 76.44 & 90.22 & 89.48 & 93.74 & 91.83 & 85.89 \\
    Lumina-Image 2.0~\citep{qin2025lumina} & 2.6B & 50 & - & 91.97 & 90.20 & 94.85 & - & 87.20 \\
    Seedream 3.0~\citep{gao2025seedream} & / & / & 94.31 & 92.65 & 91.36 & 92.78 & 88.24 & 88.27 \\
    GPT Image 1 [High]~\citep{gptimage} & / & / & 88.89 & 88.94 & 89.84 & 92.63 & 90.96 & 85.15 \\
    \midrule
    \multicolumn{9}{c}{\textit{Unified Models}} \\
    \midrule
    MetaQuery-L~\citep{pan2025transfer} & 3B & 30 & - & - & - & - & - & 81.10 \\
    BLIP3-o-8B~\citep{chen2025blip3} & 8B & 30 & - & - & - & - & - & 80.73 \\
    OpenUni-B-512~\citep{wu2025openuni} & 1.6B & 20 & 85.87 & 87.33 & 86.54 & 86.91 & 89.43 & 80.29 \\
    Tar-7B~\citep{han2025vision} & 9.6B & 50 & - & 88.62 & 88.05 & 93.98 & - & 84.19 \\
    TBAC-UniImage-3B~\cite{xu2025tbac} & 4.6B & 30 & 83.52 & 87.94 & 87.80 & 87.17 & 87.02 & 80.97 \\
    Qwen-Image~\citep{wu2025qwen} & 20B & 50 & 91.32 & 91.56 & 92.02 & 94.31 & 92.73 & 88.32 \\
    \midrule
    \multicolumn{9}{c}{\textit{Distilled Models}} \\
    \midrule
    SDXL-DMD2~\citep{yin2024improved} & 2.6B & 4 & 81.16 & 80.68 & 82.47 & 83.52 & 80.05 & 74.24 \\
    SD3.5-L-Turbo~\citep{esser2024scaling} & 8B & 4 & 90.99 & 87.43 & 87.42 & 87.81 & 86.10 & 81.97 \\
    SD3.5-Turbo~\cite{sauer2024fast} & 8B & 4 & 80.12 & 86.13 & 84.73 & 91.86 & 78.29 & 79.03 \\
    FLUX.1-schnell~\citep{flux2024} & 12B & 4 & 86.62 & 90.82 & 88.35 & 93.45 & 82.00 & 84.94 \\
    SANA-Sprint~\citep{chen2025sana} & 1.6B & 4 & 83.84 & 88.54 & 88.50 & 87.40 & 86.41 & 81.08 \\
    \midrule
    \multicolumn{9}{c}{\textit{BLIP3o-NEXT and Ours under Few-Step Generation}} \\
    \midrule
    \multirow{4}{*}{BLIP3o-NEXT} & 3B & 1 & 73.60 & 69.10 & 73.48 & 79.92 & 69.09 & 57.05 \\
     & 3B & 2 & 82.32 & 79.35 & 79.16 & 77.71 & 79.66 & 67.38 \\
     & 3B & 4 & 88.53 & 85.99 & 85.04 & 87.78 & 86.44 & 78.15 \\
     & 3B & 30 & 86.21 & 88.55 & 86.82 & 90.14 & 88.01 & 82.05 \\
    \cmidrule(lr){1-9}
    \multirow{4}{*}{EMF} & 3B & 1 & 85.24 & 85.85 & 85.19 & 82.37 & 82.65 & 77.36 \\
     & 3B & 2 & 85.63 & 88.15 & 85.96 & 85.69 & 86.20 & 79.44 \\
     & 3B & 4 & 88.01 & 87.27 & 88.24 & 88.78 & 87.68 & 81.20 \\
     & 3B & 8 & 89.07 & 88.13 & 88.96 & 87.49 & 86.34 & 81.94 \\
    \bottomrule
    \end{tabular}
    % }
\end{table*}

\noindent\textbf{Instruction Complexity}. Class labels typically encapsulate a single semantic concept, whereas natural language prompts often bind multiple attributes, objects, and spatial relations (e.g., a blue ceramic teapot on a wooden table next to a vase of red tulips). In few-step regimes, the model has limited opportunities for correction. Therefore, inadequate \emph{disentanglement} of these semantic components can easily lead to binding errors, missing objects, or incorrect attribute assignments.

The generation dynamics differ significantly between class-label and textual conditioning, a contrast visualized in Fig.~\ref{fig:v_field}. Under the simpler class-label conditioning, the denoising trajectory is relatively smooth. This smoothness indicates that the instantaneous velocity at each step closely aligns with the overall average velocity, making it straightforward for the model to predict this average. This stability is rooted in the embedding space, where class-label features form sparse clusters with large inter-class margins, ensuring category integrity and attribute accuracy even in single-step generation.

In stark contrast, the higher complexity and coupled nature of textual conditions lead to a more tortuous denoising trajectory. This winding path causes a significant divergence between the instantaneous and average velocities, often manifesting as early-stage semantic drift. Consequently, the model struggles to converge on the correct average velocity, necessitating additional corrective steps. This difficulty is exacerbated by the nature of textual embeddings, which reside in densely packed neighborhoods and inherently complicate the estimation of a stable average velocity.

These observations directly link the challenges of text-conditioned MeanFlow to the key properties of high-quality textual representations introduced in the main text: strong \emph{discriminability} and \emph{disentanglement} are essential for preserving semantic fidelity when the velocity field is learned under limited denoising steps.

\section{Additional Experiment on text encoder}
We conducted analyses of the post-trained SANA-1.5 and OpenUni text encoders, and ran mean-flow experiments on OpenUni. We chose OpenUni because it shares the SANA-1.5 diffusion backbone, but uses a InternVL3–based text encoder.
Tab.~\ref{tab:rebuttal} compares the two encoders. After training, Gemma becomes less discriminative but more disentangled, which helps 20-step generation by refining the language space. In contrast, mean-flow few-step generation needs strong image–text discriminability, so it still fails even after encoder training.
We also train mean flow on OpenUni under the same setup (Tab.~\ref{tab:openuni-experiment}). OpenUni performs better than SANA-1.5, benefiting from stronger text-encoder representations, but it still falls short of the original model due to insufficient discriminability.

\begin{table}[!htbp]
    \centering
    \caption{Experiments on discriminability and disentanglement metrics for the trained SANA-1.5 and OpenUni text encoders.}
    \label{tab:rebuttal}
    \renewcommand{\arraystretch}{1.0}
    
    \begin{tabular}{lc}
        \toprule
        \textbf{Metric} & \textbf{Value} \\
        \midrule
        Disc. (Gemma-train)   & 0.694 \\
        Disc. (OpenUni)       & 0.724 \\
        Dise. (Gemma-train) & 0.997 \\
        Dise. (OpenUni)     & 0.996 \\
        \bottomrule
    \end{tabular}
\end{table}

% 第二个表格：OpenUni Experiment
\begin{table}[!htbp]
    \centering
    \caption{Results of OpenUni trained on Mean Flow.}
    \label{tab:openuni-experiment}
    
    \renewcommand{\arraystretch}{1.0}
    
    \begin{tabular}{ccc}
        \toprule
        \textbf{Steps} & \textbf{FM-GenEval} & \textbf{MF-Geneval} \\
        \midrule
        20 & 0.86 & 0.76 \\
        4  & 0.73 & 0.70 \\
        2  & 0.31 & 0.61 \\
        1  & 0.11 & 0.59 \\
        \bottomrule
    \end{tabular}
\end{table}

\section{Inference Time Comparison.}
When generating images from the same prompt and timing diffusion sampling only, BLIP3o-NEXT on H200 takes 1.24 s with 30 steps, while ours takes 0.22/0.12/0.08 s (4/2/1 steps). 
For end-to-end generation with different prompts, BLIP3o-NEXT (30 steps) takes 11.3 s, whereas our 4-step version takes 9.87 s. The remaining time is mostly spent on autoregressive text-embedding generation.

\section{User Study and ImageReward Result}
Considering instruction-following ability, we conducted PickScore and a user study on 50 prompts (similar to Fig.1 in our manuscript). We recruited 20 users, who compared images generated by five models for each prompt and answered: “Which result best matches the prompt?”
\begin{table}[htbp!]
\centering
\caption{Performance comparison across different models.}
\label{tab:performance-comparison}
\renewcommand{\arraystretch}{1.0} 
\begin{tabular}{lcc}
\toprule
\textbf{Model} & \textbf{PickScore} & \textbf{User Study} \\
\midrule
SDXL-DMD2      & 0.14 & 0.09 \\
SD3.5-L-Turbo  & 0.16 & 0.13 \\
FLUX.1-schnell & 0.17 & 0.12 \\
SANA-Sprint    & 0.25 & 0.16 \\
Ours           & \textbf{0.28} & \textbf{0.49} \\
\bottomrule
\end{tabular}
\end{table}

All models use 4-step generation, and both experiments show that our method performs better.

\section{Additional Quantitative and Qualitative Results}
\label{sec:Additional Visual Results}

We provide supplementary quantitative and qualitative evaluations to further validate the effectiveness of our approach under limited denoising steps.

\textbf{DPG-Bench evaluation.}
Generating high-fidelity images from complex and detail-rich textual prompts in a limited number of denoising iterations is a highly challenging task. To assess our model's capability in this regime, we conduct extensive tests on DPG-Bench, which focuses on long-form prompts with intricate attribute bindings and spatial relationships. As reported in Tab.~\ref{tab:dpg}, our method consistently outperforms equally sized distilled few-step models under the same step setting, despite the inherent difficulty of the benchmark. Notably, with only 8 sampling steps, our model delivers performance on par with the BLIP3o-NEXT baseline using 30 steps, and even under the challenging \emph{1-step} regime, it surpasses widely-used distilled models such as SDXL-DMD2 and Playground~v2.5 in overall score.

\textbf{Vertical comparison across sampling steps.}
We additionally present the few-step generation results of our MeanFlow adaptation under \emph{1-step}, \emph{2-step}, \emph{4-step}, and \emph{8-step} settings, comparing them with the BLIP3o-NEXT baseline trained with standard Flow Matching under the same sampling step configurations. 

\begin{figure*}[b!]
    \centering
    \includegraphics[width=1.0\linewidth]{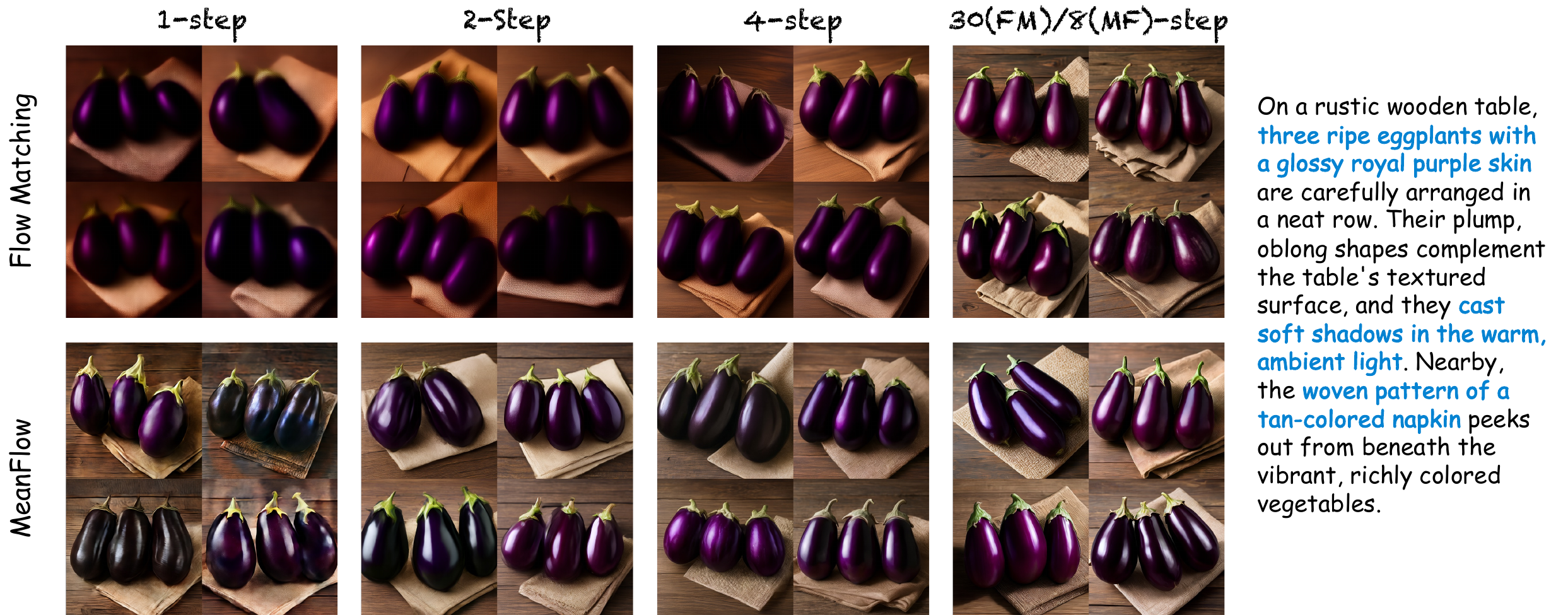}
    % \vspace{-2.0em}
    \caption{Representative visual results on DPG-Bench. Compared to the blurred outputs of few-step Flow Matching~(FM) inference, our MeanFlow~(MF) approach produces relatively sharp images even with a single sampling step, and with 8 sampling steps achieves visual quality comparable to Flow Matching using 30 steps, demonstrating a favorable trade-off between generation speed and visual fidelity.}
    \label{fig:vis compare fm or mf}
    % \vspace{-1em}
\end{figure*}

As shown in Fig.~\ref{fig:vis compare fm or mf}, our method achieves an effective trade-off between inference speed and output quality: 
whereas the Flow Matching baseline exhibits noticeable blurring and loss of fine details when the number of sampling steps is reduced, our MeanFlow sampling retains salient object structures and fine-grained textures even at extremely low step counts, producing visually coherent and semantically faithful images at a fraction of the baseline’s inference time.

\begin{figure*}[b!]
    \centering
    \includegraphics[width=1.0\linewidth]{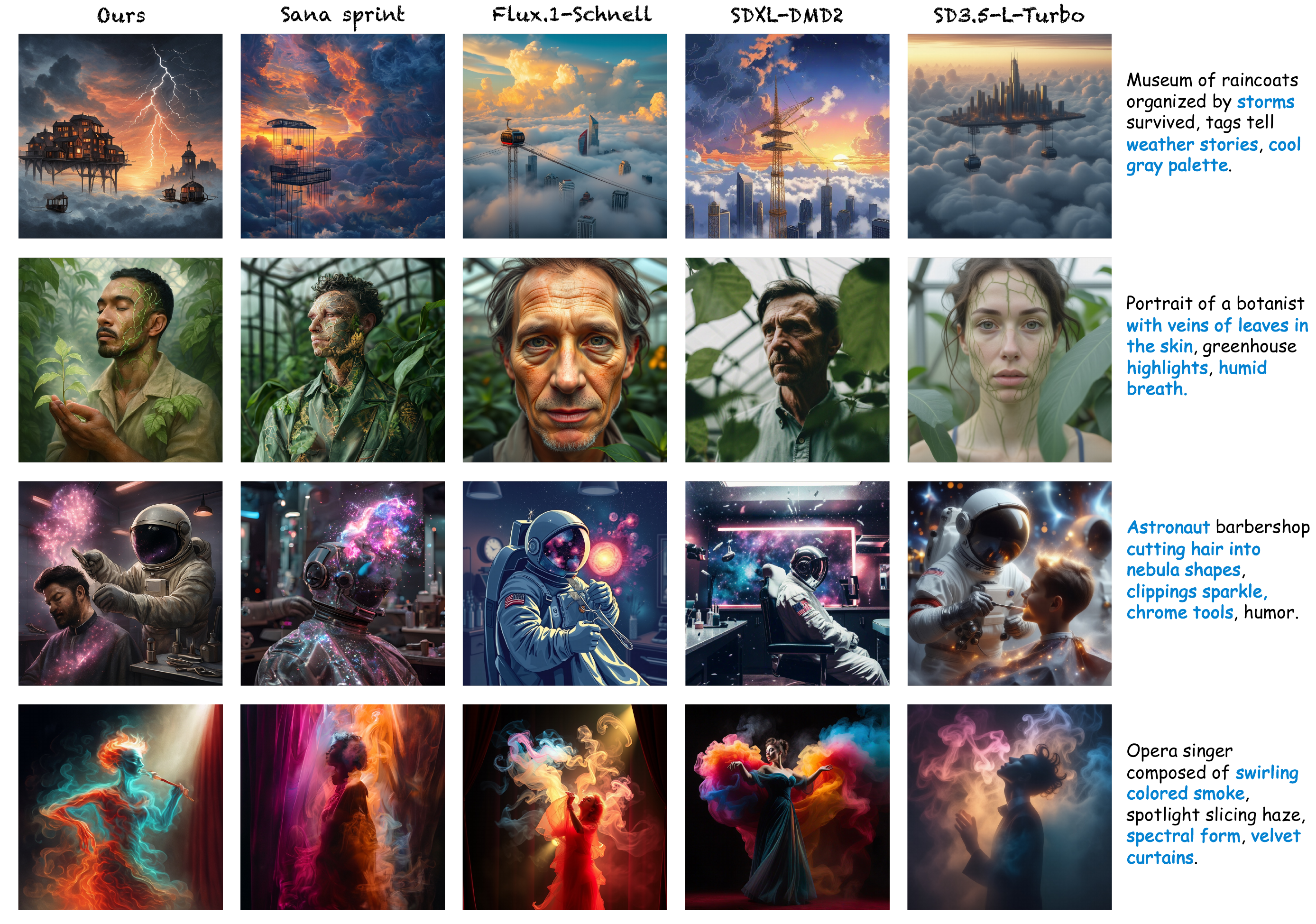}
    % \vspace{-2.0em}
    \caption{Additional comparisons under 4-step sampling between our method and existing distilled models. Our approach achieves higher semantic fidelity and richer visual details, closely adhering to complex text prompts. Blue text indicates cases where competing models fail to accurately render the described content.}
    \label{fig:vis compare 4 step}
    % \vspace{-1em}
\end{figure*}

\textbf{Horizontal few-step comparison.}
We also present side-by-side comparisons between our model and other few-step approaches at same 4-step settings (Fig.~\ref{fig:vis compare 4 step}). These results highlight our model's ability to preserve fine-grained details and adhere to textual instructions more faithfully than existing distilled models, across a diverse set of challenging prompts.

\clearpage

\newpage

\clearpage

% WARNING: do not forget to delete the supplementary pages from your submission 
% \input{sec/7_appendix}

\end{document}